\documentclass[format=sigconf, review=false, anonymous=false, authorversion=true]{acmart}

\usepackage{amssymb,amsmath,amsthm}
\usepackage{algorithm}
\usepackage{algorithmicx}
\usepackage{algpseudocode}
\usepackage{dsfont}
\usepackage{multirow}
\usepackage {subcaption} 

\makeatletter
\DeclareRobustCommand\onedot{\futurelet\@let@token\@onedot}
\def\@onedot{\ifx\@let@token.\else.\null\fi\xspace}

\AtBeginDocument{%
  \providecommand\BibTeX{{%
    \normalfont B\kern-0.5em{\scshape i\kern-0.25em b}\kern-0.8em\TeX}}}

\acmYear{2023}\copyrightyear{2023}
\setcopyright{rightsretained}
\acmConference[MMSys '23]{Proceedings of the 14th ACM Multimedia Systems Conference}{June 7--10, 2023}{Vancouver, BC, Canada}
\acmBooktitle{Proceedings of the 14th ACM Multimedia Systems Conference (MMSys '23), June 7--10, 2023, Vancouver, BC, Canada}
\acmPrice{15.00}
\acmDOI{10.1145/3587819.3590988}
\acmISBN{979-8-4007-0148-1/23/06}

\begin{document}

\title{Video-based Contrastive Learning on Decision Trees:  from Action Recognition to Autism Diagnosis}

\author{Mindi Ruan}
\email{mr0114@mix.wvu.edu}
\affiliation{%
\institution{Lane Department of Computer Science and Electrical Engineering, West Virginia University}
 \streetaddress{PO Box 6109}
 \city{Morgantown, WV}
 \country{USA}}

\author{Xiangxu Yu}
\email{xiangxu@wustl.edu}
\affiliation{%
 \institution{Department of Radiology, Washington University  in St. Louis}
 \streetaddress{Campus Box 8225, 4525 Scott Ave}
 \city{St. Louis, MO 63110}
 \country{USA}}
 
\author{Na Zhang}
\email{nz0001@mix.wvu.edu}
\affiliation{%
\institution{Lane Department of Computer Science and Electrical Engineering, West Virginia University}
 \streetaddress{PO Box 6109}
 \city{Morgantown, WV}
 \country{USA}}
 
\author{Chuanbo Hu}
\email{chuanbo.hu@mail.wvu.edu}
\affiliation{%
\institution{Lane Department of Computer Science and Electrical Engineering, West Virginia University}
 \streetaddress{PO Box 6109}
 \city{Morgantown, WV}
 \country{USA}}
  
\author{Shuo Wang}
\email{shuowang@wustl.edu}
\affiliation{%
 \institution{Department of Radiology, Washington University  in St. Louis}
 \streetaddress{Campus Box 8225, 4525 Scott Ave}
 \city{St. Louis, MO 63110}
 \country{USA}}

\author{Xin Li}
\authornote{Corresponding author}
\email{xin.li@mail.wvu.edu}
\affiliation{%
 \institution{Lane Department of Computer Science and Electrical Engineering, West Virginia University}
 \streetaddress{PO Box 6109}
 \city{Morgantown, WV}
 \country{USA}}

\begin{abstract}
How can we teach a computer to recognize 10,000 different actions? Deep learning has evolved from supervised and unsupervised to self-supervised approaches. In this paper, we present a new contrastive learning-based framework for decision tree-based classification of actions, including human-human interactions (HHI) and human-object interactions (HOI). The key idea is to translate the original multi-class action recognition into a series of binary classification tasks on a pre-constructed decision tree. Under the new framework of contrastive learning, we present the design of an interaction adjacent matrix (IAM) with skeleton graphs as the backbone for modeling various action-related attributes such as periodicity and symmetry. Through the construction of various pretext tasks, we obtain a series of binary classification nodes on the decision tree that can be combined to support higher-level recognition tasks. Experimental justification for the potential of our approach in real-world applications ranges from interaction recognition to symmetry detection. In particular, we have demonstrated the promising performance of video-based autism spectrum disorder (ASD) diagnosis on the CalTech interview video database.
\end{abstract}





\maketitle
\section{Introduction}
\label{sec:intro}
Rapid advances in smartphone technology have greatly facilitated the acquisition and sharing of short video clips through various social media platforms such as TikTok and Instagram. Video-based action recognition \cite{ji20123d,wang2013action,feichtenhofer2016convolutional,carreira2017quo,zhu2020comprehensive} has been extensively studied in the literature. In contrast, the recognition of human interaction based on video \cite{shu2019hierarchical} has remained an under-researched topic. One of the emerging applications for analyzing human interaction from short video clips is behavioral imaging \cite{rehg2014behavioral}, which aims at diagnosing behavioral disorders (e.g., autism and depression) from short video clips in natural environments. In particular, the online diagnosis of autism spectrum disorder (ASD) at home \cite{abbas2018machine} has received increasingly more attention from both the technical and clinical communities. During the COVID pandemic, the conventional gold standard, such as the Autism Diagnostic Observation Schedule (ADOS) interview \cite{lord2000autism}, has become unsafe to practice due to the regulation of social distancing. Instead, behavioral imaging of short video clips acquired by smartphones becomes a safer and more cost-effective alternative to ADOS interviews.

From the multimedia perspective, the key challenge in action recognition lies in the complexity of human movements and social interactions. Similarly, the diagnosis of ASD from behavioral imaging also faces challenges in the automatic analysis of social affection (SA) and restricted and repetitive behavior (RRB) from smartphone video. Both the human-human interaction (HHI) \cite{perez2021interaction} and the human-object interaction (HOI) \cite{yao2010modeling} are important to the task of behavioral imaging: the former reflects the subject's social interaction skills, and the latter is correlated with the person's motor as well as joint attention skills. 
In behavioral imaging, abnormal behaviors in HHI and HOI often carry significant information to the diagnosis of autism, e.g., lack of response to name calling \cite{campbell2019computer} and sticking to the same repeating pattern during toy play \cite{toth2006early}. Other abnormal behaviors not involved with interaction, such as avoiding eye contact during conversation \cite{jiang2017learning} and self-stimulating called stimming \cite{rajagopalan2013self} are also hallmark signals for autistic children.

Although both HHI and HOI have been studied for action recognition by computer vision and multimedia communities, irregularity detection \cite{boiman2007detecting} or anomaly detection \cite{georgescu2021anomaly} from video clips has several unique challenges. In addition to well-known barriers (e.g., cluttered background, interference from occlusion, and invariance to viewpoint \cite{kong2018human}), it is often difficult for even human experts to tell the subtle differences between regular hand flapping and stimming or between toy playing by a typically developing (TD) and the sameness (sticking to the same toy - a typical repetitive behavior) by autistic children. For example, an hour-long ADOS interview video, when viewed by non-trained eyes, often fails to distinguish between TD and ASD most of the time. It takes special training for health professionals, such as developmental pediatricians and behavioral therapists, to become ADOS-reliable. Teaching a computer to identify irregularities (e.g., abnormal behavior in autistic patients) from video reliably has remained a long-standing open problem in computer vision \cite{georgescu2021anomaly}.

Inspired by the long-lasting impact of classification and regression trees (CART) \cite{breiman2017classification}, we propose to take a binary decision tree approach to address both action recognition and diagnosis of ASD in this work. New insights are borrowed from human motion analysis \cite{aggarwal1999human} and autism research \cite{lord2000autism} to focus on a series of basic binary classification tasks as the building block of the complete vision system. For action recognition, we have considered mutual interaction (absent vs. present), time reversibility (yes vs. no), periodicity (yes vs. no) and body movement (upper vs. lower). For the diagnosis of ASD, we have focused on modeling two types of irregularities associated with autistic behavior: $symmetry$ breaking social interaction \cite{williams2001imitation} (e.g. lack of social interaction and imitation) and sticking to the $sameness$ \cite{gotham2013exploring} (e.g., repetitive actions and self-stimulating behaviors). The shared motivation behind our approach to naive Bayesian decision trees \cite{zadrozny2001obtaining} offers a unified solution to both action recognition and ASD diagnosis, i.e., for a given video clip, our objective is to produce a sequence of binary decisions constructed from domain knowledge. By fusing those binary decisions adaptively, we expect to improve the reliability and accuracy of video-based action recognition or diagnosis of ASD. 

To support inference in the decision tree, we propose to construct an interaction adjacency matrix (IAM) on skeleton graphs \cite{yan2018spatial} for representation learning and develop self-supervised contrastive learning \cite{lin2020ms2l} for binary classification. Our IAM representation can be interpreted as an extension of the graph convolution network (GCN) \cite{liu2020disentangling} by incorporating HHI into the adjacency matrix. Our IAM is conceptually similar to recent pairwise adjacency matrix (PAM) \cite{yang2020pairwise} and interaction relational network (IRN) \cite{perez2021interaction} for HHI detection. However, PAM lacks interaction information from one part of a person to a different part of another person, as it only focuses on identical joint types or the center of gravity interaction between two individuals. This limitation is particularly noticeable in asymmetric mutual actions, such as when one person (dominant) punches another person's face (subordinate). In such a case, the dominant person's hand should exhibit a strong intensity towards the subordinate's face in the reconstructed interaction graph. Our IAM overcomes this limitation and, unlike IRN, eliminates the need for additional computing resources for a relation network \cite{santoro2017simple} to compute the relationships between joint pairs. Moreover, our IAM contains a novel extension, i.e., the interaction between the left and right hands of a single person. Meanwhile, the new insight brought about by knowledge of the autism domain allows us to learn the concept of symmetry in interaction and imitation as a behavioral marker, i.e., autistic people often demonstrate broken symmetry in both HHI (e.g., lack of response to name-calling \cite{liu2017response}) and HOI imitation (e.g., pretending to brush the teeth during the ADOS interview \cite{smith1994imitation}).
Our technical contributions can be summarized as follows.

\begin{itemize}
\item Formulation of the problem of video-based action recognition and diagnosis of ASD via decision trees. 
    We aim to extract a series of basic units, such as periodicity and dominance, for action recognition and behavioral biomarkers (symmetry and sameness) to facilitate the task of detecting ASD.
    
\item Construction of a trainable ST-GCN for representation learning. We propose to learn an IAM for modeling the interaction between two people and two hands of a single person. Instead of using a learnable mask \cite{yan2018spatial}, we have developed a different IAM as a warm-up regularization strategy.
    
\item Development of a self-supervised contrastive learning algorithm for attribute/anomaly detection. We have built on previous work \cite{lin2020ms2l} to tackle the problem of periodicity/dominan-ce/symmetry/sameness detection by constructing pretext tasks. By combining ST-GCN representation with contrastive learning, we managed to achieve a reliable and transparent extraction of features from the action video and biomarkers from the ADOS video. The excellent explainability of our approach is desirable for bridging between computing and healthcare professionals.

\item Experimental results are reported to demonstrate the performance of our approach in video-based action recognition and diagnosis of ASD. On action recognition, we have achieved a noticeable improvement over previous ST-GCN on several testing scenarios. Using symmetry and similarity as biomarkers, respectively, we have achieved the precision of $>80\%$ for the task of diagnosing ASD. 
\end{itemize}

\section{Related Work}  
\label{sec:related}
\textbf{Video-based action Recognition}. Previous work on action recognition is based on 3D convolutional neural networks \cite{ji20123d} or two-stream convolutional networks \cite{simonyan2014two}. The experiments are often conducted on two popular datasets: HMDB-51 \cite{Kuehne11} and UCF-101 \cite{soomro2012ucf101}. In view of the paucity of videos in these datasets, a new Kinetics Human Action Video dataset \cite{kay2017kinetics} was used in \cite{carreira2017quo} along with a new two-stream inflatable 3D ConvNet (I3D) architecture. More recently, the construction of larger datasets for action recognition, such as NTU RGB+D \cite{shahroudy2016ntu} and NTU RGB+D 120 \cite{liu2019ntu} has inspired a flurry of more powerful architectures (e.g., graph convolutional network \cite{yan2018spatial,li2019actional,cheng2020skeleton}, transformer \cite{plizzari2021spatial,zhang2021stst,li2021trear}) for action recognition. The accuracy of current state-of-the-art skeleton-based action recognition has exceeded $76\%$ and $91\%$ in Kinetics-400 and NTU RGB+D, respectively. There is also a recent study on group activity recognition \cite{wu2019learning}, which emphasizes the group instead of individual actions, as well as symmetry modeling for action evaluation \cite{AIM} through an asymmetric interaction module. For a recent survey on video action recognition, see \cite{zhu2020comprehensive}. The latest advances in this field include recurring the transformer \cite{yang2022recurring}, dual-head contrastive domain adaptation \cite{da2022dual}, Direcformer \cite{truong2022direcformer}, and learning from the temporal gradient \cite{xiao2022learning}.


\noindent \textbf{Video-based ASD Classification}. In the literature on autism research, video-based autism diagnosis dated back at least to \cite{ozonoff2011onset} where the correspondence between home video and parent report was shown as an onset pattern in autism. Early works on behavioral imaging for autism screening \cite{rehg2014behavioral} focused on abnormal eye contact or visual scanning of faces \cite{pelphrey2002visual}. Eye-tracking data for visual attention have been used to train a deep neural network (DNN) for identifying people with ASD in \cite{jiang2017learning}. The idea of ASD screening using machine learning on home video has also been explored in \cite{abbas2018machine} and \cite{tariq2018mobile}; however, the extraction and scoring of 30 diagnostic features (e.g., eye contact, joint attention, and imitate actions) was still done by humans in \cite{tariq2018mobile} due to the technical challenge with automatic video analysis. Most recently, the gaze pattern extracted from the video has been utilized for early ASD diagnosis of young children in  \cite{chang2021computational}. Generally speaking, there is still a significant gap between human-based and machine-based analysis of ADOS interview video \cite{lord2000autism} for the autism diagnosis. When compared with standard HHI and HOI research \cite{perez2021interaction,yang2020pairwise,GeomNet,hou2020visual,zhou2020cascaded,wang2020learning}, video-based ASD classification is lacking in both the dataset and pre-trained models. Only recently, video-based ASD diagnosis has been studied for remote telehealth assessment in \cite{dahiya2020systematic} and with few-shot learning constraints in \cite{zhang2022discriminative}.


\section{The Proposed Method}
\label{sec:method}
\subsection{Problem Formulation and Preliminaries}

\begin{figure}[h]
    \centering
    \includegraphics[width=\columnwidth,height=1.8in]{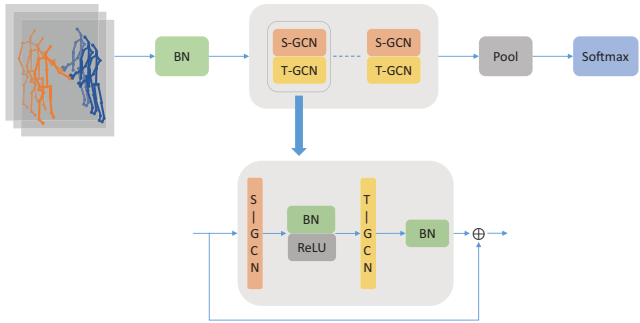}
    \caption{The architecture of our backbone: spatiotemporal graph convolutional network (ST-GCN) \cite{yan2018spatial}.}
    %
    \label{fig:architecture}
\end{figure}

\begin{figure*}[htb]
    \centering
    \vspace{-0.05in}
    \includegraphics[width=1.8\columnwidth]{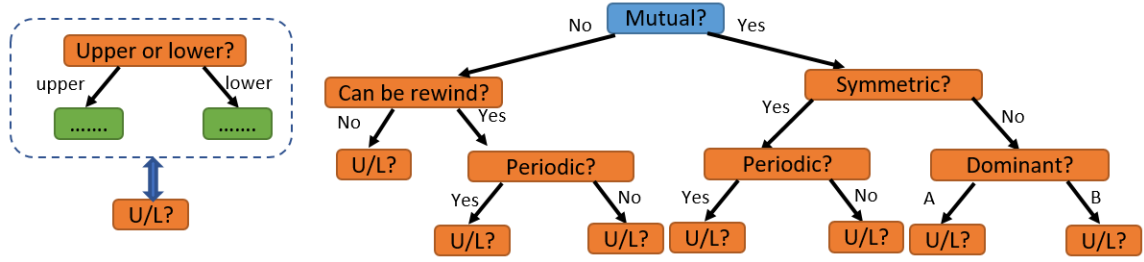}
    \caption{Extending binary decision tree for multi-class action recognition (note that an $N$-class action recognition can be decomposed into a binary decision tree with a height of approximately $log_2 N$).}
    %
    \label{fig:decision}
\end{figure*}

In the literature, video-based action recognition refers to the problem of labeling video sequences with action labels \cite{poppe2010survey}. Unlike multi-label classification\cite{hu2021detection}, we opt to formulate the video-based action recognition and diagnosis of ASD as a binary classification problem in decision trees (as shown in Figure \ref{fig:decision} and Figure \ref{fig:intro}) for the sake of explicitly exploiting the domain knowledge related to SA and RRB. For a given video clip, our aim is to generate a sequence of binary scores\footnote{Note that the binary decision tree can be generalized to a quad-tree for supporting the scoring of multiple choices such as 0-3 as adopted by the previous work \cite{tariq2018mobile}.} as the output of the decision tree. This binary score vector can be used to train various machine learning models (e.g., support vector machine and random forest \cite{breiman2001random}) for the action recognition or ASD diagnosis task. With the introduction of decision trees, we start with the formulation of a binary classification problem as the building block of our recognition/diagnosis system.

\textbf{Problem Formulation}. Given a segment of an action or ADOS interview video (assuming that it is parsed into an individual interview task after preprocessing), how to determine its binary action-related or ASD-related attribute?

To tackle the above problem, we will start with a brief review of the existing work on skeleton graph representation that has been widely studied for action recognition. A skeleton graph for action recognition is an abstraction of 17 or 25 joints based on human anatomy. The current state-of-the-art in skeleton graph-based action recognition has adopted a spatiotemporal graph convolutional network (ST-GCN) \cite{kipf2016GCN} architecture. However, datasets for action recognition, such as NTU-60 and NTU-120, only contain a small collection of actions from our daily lives and are only appropriate for supervised learning. We need to extend both data representation and network design for autism diagnosis by incorporating domain-specific knowledge. 

\textbf{Skeleton Graph and ST-GCN}. Skeleton-based action recognition has received increasingly more attention thanks to the rapid advance of 3D pose estimation \cite{tome2017lifting} and graph convolutional networks (GCN) \cite{kipf2016GCN}. In this paper, we choose spatial-temporal graph convolutional networks (ST-GCN) \cite{yan2018spatial} as the baseline because it has been widely studied in the literature on action recognition. The basic idea behind ST-GCN is to construct a spatial-temporal graph based on physical connections of human joints in consecutive frames. In this graph, the baseline model used nine ST-GCN blocks to aggregate information on different scales. 

\subsection{Interaction Adjacency Matrix (IAM)}
\label{sec:IAM}

Here, we first review the basic concept of ST-GCN construction and then introduce an interaction adjacency matrix (IAM) as the new building block for modeling the imitation of HHI and HOI.
Following the notation used in \cite{yan2018spatial}, we can construct a spatial-temporal graph $G = (V,E)$ from a skeleton sequence with $N$ joints and $T$ frames (the output of the 3D pose estimation from the input video sequence) as follows. In this graph, the set of vertex $V=\{v_{ti}|t=1,...,T; i=1,...,N\}$ includes all body joints (indexed by $i$) collected together with a skeleton sequence (indexed by $t$). The graph includes two types of edges: 1) the {\em spatial} edge connects two joints within the same frame based on the human body structure; 2) the {\em temporal} edge connects the same joint in consecutive frames. Formally, the spatial edge connection can be denoted as $E_{S}=\left\{v_{t i} v_{t j} \mid(i, j) \in H\right\}$, where $H$ is the set of physical connected joints; and the temporal edge connection can be denoted as $E_{T}=\left\{v_{t i} v_{(t+1) i}\right\}$.

Note that the spatial edge connections are kept the same for all frames. In other words, we only need to construct one spatial graph for the entire sequence. Similarly to the original formulation of GCN \cite{kipf2016GCN}, the spatial graph can be directly characterized by an intraperson adjacency matrix $\mathbf{A} \in \mathbb{R}^{N \times N}$, where $\mathbf{A}_{ij}$ $(i \neq j)$ denotes the intensity of the joint $V_i$ to the joint $V_j$, or the intensity of the edge $E_{ij}$. Like an attention mark for image recognition,
the adjacency matrix serves as an adjustable kernel to embed action-related information during training. Recent GCN-based approaches \cite{cheng2020decoupling,shi2019two,shi2019skeleton} have explored the modality of a trainable intraperson adjacency matrix for action recognition. We leverage this idea to flexibly extract salient features for symmetry detection (e.g., whether both hands of a person are moving during the interview) as follows. 

An ST-GCN block consists of a GCN module and a standard convolutional module, where the GCN module plays an important role in extracting spatial-temporal information. 
Let $\mathbf{f}_{in} \in \mathbb{R}^{N \times C}$ and $\mathbf{f}_{out} \in \mathbb{R}^{N \times C^{\prime}}$ be the input and output features of the current layer, respectively, where $N$ is the number of joints in a skeleton graph. $C$ and $C^{\prime}$ are the dimensions of the input and output features separately. 
In \cite{yan2018spatial}, the skeleton graph is represented by an adjacency matrix $\mathbf{A} \in \mathbb{R}^{3 \times N \times N}$ with spatial configuration partitioning. The adjacency matrix can be further disassembled into three matrices $\mathbf{A}_{sub} \in \mathbf{R}^{N \times N}$, where $sub \in$ \{root node, centripetal group, centrifugal group\} \cite{yan2018spatial}. Then, the GCN module can be expressed as 

\begin{equation}
 \mathbf{f}_{out}=\sum_{sub} \widetilde{\mathbf{A}_{sub}} \mathbf{f}_{in} \mathbf{W}_{sub}
\label{equ:1}
\end{equation}
where $\widetilde{\mathbf{A}_{sub}} = \boldsymbol{D}_{sub}^{-\frac{1}{2}} (\mathbf{A}_{sub} + \mathbf{I}) \boldsymbol{D}_{sub}^{-\frac{1}{2}}$ is the normalized adjacency matrix; and 
$\boldsymbol{D}_{sub}^{ii} = \sum_{k}\left(A_{sub}^{i k}\right)+\alpha$, where $\alpha$ is often set to a small positive real (e.g., $0.001$) to avoid empty rows in $\mathbf{A}_{sub}$. 
The learning of the trainable weight $\mathbf{W}_{sub} \in \mathbb{R}^{C \times C^{\prime}}$ can be implemented by adding a learnable mask $\textbf{M}$ to each ST-GCN layer. The mask plays the role of scaling the contribution of a joint's features to its neighboring joints. For each adjacency matrix, we accompany it with a learnable weight matrix $\textbf{M}$ and substitute $\mathbf{A}_{sub} + \mathbf{I}$ in $\widetilde{\mathbf{A}_{sub}}$ by $(\mathbf{A}_{sub} + \mathbf{I})\otimes \textbf{M}$, where $\otimes$ denotes the product of the matrix by element.

Recently developed GCN-based methods \cite{cheng2020decoupling,shi2019two,shi2019skeleton} explored the modality of the adaptive adjacency matrix to improve the performance on action recognition. However, most of them focus only on the intraperson relationship, ignoring the mutual interaction among people. This deficiency becomes more serious for the recognition of mutual actions between two people, such as the UT interaction dataset \cite{ryoo2009spatio}. To solve this problem, ST-GCN-PAM \cite{yang2020pairwise} introduced the inter-person correlation of the same joint between two people with a pairwise adjacency matrix (PAM). Despite its conceptual simplicity, correlation-based modeling of HHI overlooks the rich and complicated dependency among humans, especially in the presence of asymmetric mutual actions (e.g., an autistic child shows no response to name-calling). 

Inspired by the need for symmetry detection for HHI, we propose an extension of the existing adjacency matrix, named the Interaction Adjacency Matrix (IAM), to effectively model the inter-person interaction and summarize both intra-joint and inter-joint relationships. The IAM is formally defined based on the original adjacency matrix of \cite{yan2018spatial} as follows.

\begin{equation}
    \mathbf{IAM}=\left[\begin{array}{ccc}
    \mathbf{A^{11}} & \mathbf{A^{12}} \\
    \mathbf{A^{21}} & \mathbf{A^{22}}
    \end{array}\right] \in \mathbb{R}^{3 \times 2N \times 2N}
    \label{eq:2}
\end{equation}
where $\mathbf{A}^{pq} \in \mathbf{R}^{N \times N}$, and $p, q \in \{1, 2\}$ denote two people in HHI or a person with his mirror symmetry in HOI imitation. When $p = q$, $\mathbf{A}^{pq}$ indicates the intra-person adjacency matrix (degenerate to the basic case $\textbf{A}_{sub}$) for HHI and HOI imitation. When $p \neq q$, we have an inter-person 
adjacency matrix characterizing the HHI between $p$ and $q$ or the left-right adjacency matrix for HOI imitation.
Without prior knowledge of the presence of any interactions, we expect the model to automatically add or remove edges during the construction of an effective graph for $\mathbf{A^{12}}$ and $\mathbf{A^{21}}$. Instead of using a learnable mask $\mathbf{M}$ in ST-GCN \cite{yan2018spatial},  we introduce a difference matrix $\mathbf{B}_{sub}$ and substitute $(\mathbf{A}_{sub} + \mathbf{I})\otimes \textbf{M}$ in $\widetilde{{\mathbf{A}}_{sub}}$ with $\mathbf{A}_{sub} + \mathbf{B}_{sub} + \mathbf{I}$. Then Eq. \eqref{equ:1} can be rewritten as

\begin{equation}
  \mathbf{f}_{out}=\sum_{sub} \widetilde{\mathbf{C}_{sub}} \mathbf{f}_{in} \mathbf{W}_{sub}
  \label{equ:3}
\end{equation}
where $\widetilde{\mathbf{C}_{sub}}=\mathbf{A}_{sub} + \mathbf{B}_{sub} + \mathbf{I}$ and $\mathbf{B}_{sub} \in \mathbf{R}^{3 \times N \times N}$ is a training matrix initialized with zeros.  In contrast, $\mathbf{A}_{sub}$ is a fixed adjacency matrix that can be interpreted as the average. Using such a decomposition of the average difference, we can easily add new edges or remove old edges by modifying the matrix $\mathbf{B}$ in the learning process. Furthermore, our IAM is constructed as a directed graph (a similar idea exists in directed graph neural networks \cite{shi2019skeleton}), where we can learn the relationship of directional interaction by comparing $\mathbf{A^{12}}$ and $\mathbf{A^{21}}$ for the detection of symmetry.

$\mathbf{A^{12}}$ and $\mathbf{A^{21}}$ can be extended to describe the relationship not only between humans and humans but also between humans and objects. In this paper, we leave the relationship between humans and objects part for future work and focus on the relationship between humans and humans. We apply the IAM mechanism to our model and improve the performance of mutual action recognition on the mutual action subset of NTU RGB+D 120 and HOI imitation tasks on the Caltech ADOS Interview dataset.


\subsection{Contrastive Learning for ST-GCN based Binary Classification}
\label{sec:3.3}

As mentioned before, video-based action recognition faces several technical challenges, from the cluttered background to the lack of anomalous events during training. Multitask self-supervised learning (M$S^2$L) \cite{lin2020ms2l,georgescu2021anomaly, hu2020classification} has emerged as a promising solution to overcome these challenges. In \cite{lin2020ms2l}, three tasks are integrated into the learning skeleton features for action recognition. In \cite{georgescu2021anomaly}, a similar construction of three self-supervised tasks (time arrow, motion irregularity, and object prediction) is combined with the distillation of knowledge for the detection of anomalies. The common principle is to first construct proxy tasks (e.g., puzzle recognition, object prediction) and then transfer the knowledge learned from self-supervised learning to downstream vision tasks (e.g., action recognition and anomaly detection).

Inspired by the success of M$S^2$L \cite{lin2020ms2l,georgescu2021anomaly}, we propose to design a hybrid approach for ST-GCN based binary classification, as shown in Figure \ref{fig:decision}. Similarly to \cite{lin2020ms2l}, our aim is to model skeleton dynamics instead of video frames through motion prediction and learning behavioral patterns by solving jigsaw puzzles. But unlike \cite{lin2020ms2l}, we redesign the tasks of motion prediction and jigsaw puzzle by borrowing the idea of irregularity detection from \cite{georgescu2021anomaly}. In the context of the diagnosis of ASD, the irregularity of behavior is specifically related to broken symmetry (in HHI and HOI imitation) or persistent sameness (in HOI). Such domain knowledge can be translated into specially designed jigsaw puzzles. For example, the sameness in HOI can be exploited to predict the time arrow at the object level \cite{wei2018learning}. Social disconnection, as reflected by the broken symmetry in HHI, can lead to the prediction of directional action \cite{li2020directional}.  Broken symmetry in HOI imitation can be interpreted as a kind of motion irregularity - the failure to predict the motion of the left hand based on that of the right hand. 

We apply ST-GCN\cite{yan2018spatial} as a backbone to the MS$^2$L\cite{lin2020ms2l} method and construct our self-supervised learning model. Similar to MS$^2$L, we pre-train our model with three proxy tasks: (A) a motion prediction task, (B) a jigsaw puzzle task, and (C) a contrastive learning task. 

In the motion prediction task, a sequence of consecutive frames is masked from the input skeleton sequence. We use an encoder-decoder to guide the model learning to predict the most likely poses of the masked frame. 
The input skeleton sequence is $\mathrm{X}^i=\left\{\mathrm{x}_1^i, \ldots, \mathrm{x}_T^i\right\}$, and the masked sequence is $\mathrm{X}_m^i=\left\{\mathrm{x}_1^i, \ldots, \mathrm{x}_{T^{\prime}}^i, T^{\prime}<T\right\}$. The predicted motion sequence is $\hat{\mathrm{X}}_m^i=\left\{\hat{\mathrm{x}}_{T^{\prime}+1}^i, \ldots, \hat{\mathrm{x}}_T^i\right\}$. $N$ is the batch size. The mean square error (MSE) is used to estimate the motion prediction loss $\mathcal{L}_{m}$ as follows:

\begin{equation}
  \mathcal{L}_m=\sum_{i=1}^N \sum_{t=T^{\prime}+1}^T\left\|\hat{\mathrm{x}}_t^i - \mathrm{x}_t^i\right\|_2^2
  \label{equ:4}
\end{equation}

In the jigsaw puzzle task, the skeleton sequences are divided into three segments that are randomly shuffled. The model is trained to predict the correct sequence order.
Using the shared encoder, we employ an MLP as a classifier. The predicted jigsaw category is $\hat{y}^i$. $y^i$ is a one-hot vector that denotes the jigsaw label. We use a cross-entropy loss for sequence order classification, the loss $\mathcal{L}_j$ as follows:

\begin{equation}
  \mathcal{L}_j=-\sum_{i=1}^N y^i \log_2 \hat{y}^i
  \label{equ:5}
\end{equation}

Finally, the masked input and the jigsaw input serve as transformation operations for contractive learning.
Let $z_1, z_2, \ldots, z_{N M}$ be the output of the encoder, for any integer $k$ from 1 to $N, \mathrm{z}_{(k-1) M+1}$ is the original data and the sequences from $\mathrm{z}_{(k-1) M+2}$ to $\mathrm{z}_{k M}$ are the transformed samples from the original sequence. $\overline{\mathrm{z}}_i=\frac{1}{M} \sum_{j=(i-1) M+1}^{i M} \mathrm{z}_j$ indicates the mean features of the original and transformed data for $z_{(i-1) M+1}$. $sim(\cdot)$ is the cosine similarity. $(M-1)$ is the number of transformation operations. Positive pairs are constructed from the transformation operation, while negative pairs are constructed with other samples. With $k = \left\lceil\frac{i}{M}\right\rceil$, the contractive loss $\mathcal{L}_c$ is formulated as follows:

\begin{equation}
  \mathcal{L}_c=-\sum_{i=1}^{M N} \log \frac{\exp \left(\operatorname{sim}\left(\mathrm{z}_i, \overline{\mathrm{z}}_k\right)\right)}{\sum_{j=1}^N \exp \left(\operatorname{sim}\left(\mathrm{z}_i, \overline{\mathrm{z}}_j\right)\right)},
  \label{equ:6}
\end{equation}

Our model is pre-trained with the above proxy tasks and then fine-tuned for the downstream binary classification. We trained our model for 80 epochs, using a batch size of 64 on two Nvidia RTX 3090 Ti GPUs. 

The binary classification includes: (A) asymmetric and symmetric mutual action recognition (AandS), (B) upper- and lower-body action recognition (UandL), and (C) periodic and aperiodic action recognition (periodicity). We define the class of symmetric interactions first. In our daily lives, when two people do the same action, like hugging, shaking hands, and high-five, the social protocol dictates the symmetry of both parties in the action space. We call these kinds of actions symmetric interaction (sym). Then we define dominance. When an initiating person's action leads to the following action of another person, like pushing, kicking, and pointing fingers, the two parties take an asymmetric (active vs. passive) role in the action space - i.e., dominance (the person initiating the interaction) and  subordination (the person receiving the interaction). We call these kinds of actions asymmetric interaction (asym).  
In the UandS task, the model is trained to classify the data into upper or lower body actions. 
The upper-body action refers to movements or actions that are primarily triggered by the upper part of the body, including the arms, head, and torso, such as brushing teeth, putting on a jacket, clapping, etc. In contrast, lower-body actions indicate those actions triggered by the hips, legs, and feet, such as standing up, kicking something, hopping, etc.
In the periodicity task, the model is trained to classify the data into periodic or aperiodic actions. The class of periodic actions refers to those actions that repeat themselves regularly over a certain period, such as brushing teeth, clapping, hand waving, etc. The class of aperiodic actions denotes those actions that do not repeat themselves in a regular or consistent pattern, such as standing up, kicking something, hopping, etc.

\begin{figure*}[htb]
\centering
\includegraphics[width=1.9\columnwidth]{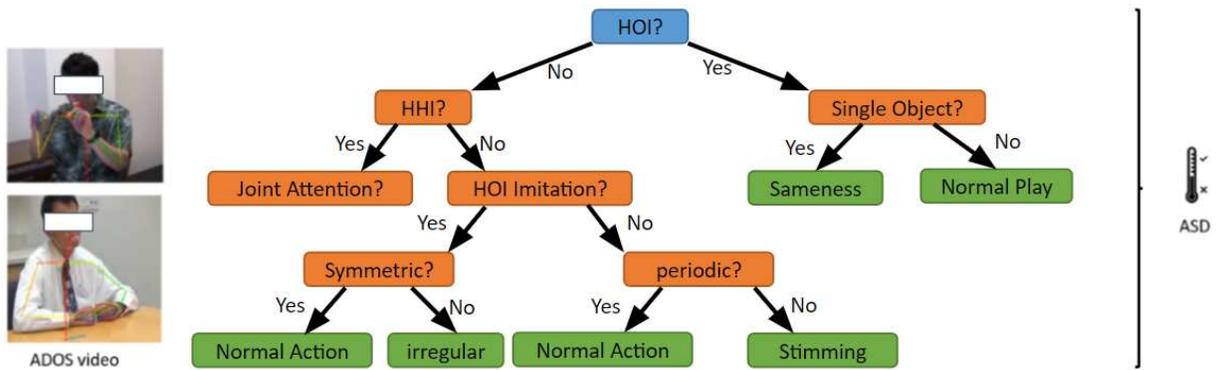}
\caption{Construction of a binary decision tree for video-based autism spectrum disorder (ASD) diagnosis. Note that we have only drawn a partial tree here for illustration purposes: we will focus on the leaf nodes of symmetry and sameness in this work (other non-leaf nodes, such as joint attention and stimming detection, are left for our future work).}
\label{fig:intro}
\end{figure*}

\section{Decision Tree for Action Recognition and ASD Diagnosis}
\label{sec:4}


\subsection{Action Recognition}

Action recognition is one of the most representative tasks for video understanding \cite{zhu2020comprehensive}. Dozens of datasets such as NTU-120 \cite{liu2019ntu}, YoutTube8M \cite{abu2016youtube}, and Kinetics-700 \cite{carreira2019short} have been constructed to support the study on video-based action recognition. Despite the tremendous progress made in the past decade, an important limitation is the lack of generalization property. Most existing action recognition methods are based on supervised learning. They cannot handle sophisticated interactions at social events. There is a need to extend the existing framework of action recognition to activity recognition by following an unsupervised and continual learning approach.

Figure \ref{fig:decision} shows the first step in this direction. Our intuition is to decompose action/activity recognition into a series of binary classification problems (the building block constructed in the previous subsection) along the pre-trained decision tree.
At the root level (marked in blue), the decision is about whether the video contains mutual action. For single-person action (left branch), we can classify the action into temporally reversible (e.g., random dancing or jumping robes) or irreversible (e.g., taking off vs. putting on the clothes). For mutual action (right branch), the interaction can be symmetric (e.g., hand-shaking or hugging) or dominant (e.g., some person pushed another). Both reversible actions and symmetric interactions can be further divided into periodic and aperiodic. At a more basic level, all actions or interactions can be classified into upper-body or lower-body movements.

We note that the construction of the above decision tree is not unique or optimal. In theory, the problem of video action/activity recognition can be solved by unsupervised online learning of decision trees in a hierarchical manner \cite{held1997unsupervised}. Unlike unsupervised clustering of face images \cite{lin2017proximity}, an optimal representation of human actions in the latent space has remained elusive. Some recent work \cite{sarfraz2021temporally} has adopted improved dense trajectory (IDT) \cite{wang2013action} and pre-trained 3000-dimensional feature vectors for unsupervised action segmentation. Our ST-GCN based representation, when combined with contrastive learning, has shown impressive performance on action recognition, as we will demonstrate in the next section. 

\subsection{ASD Diagnosis}
\label{sec:ASD}
The practical problem of ASD diagnosis can boil down to a series of binary decisions along the hierarchy, as shown in Figure \ref{fig:intro}. At the root level, lies the binary detection of human-object interaction (HOI). Based on the presence/absence of HOI, HHI and object detection are the next branching points on the tree. Then the class of HHI can be further divided into joint attention or HOI imitation, and the HOI can be further divided into repetitive behavior or normal play. To our knowledge, this is the first work to define the properties of {\em symmetry} and {\em sameness} to model the repetitive behavior of children with ASD.

Detecting symmetry in social interactions is motivated by several problems in computational social science \cite{lazer2009life}. Detecting dominance in social interaction \cite{rienks2005dominance} or the lack of motivation to participate in social interaction \cite{rosa1979incipient,ruan2021deep} often has direct implications for the modeling of human interactions \cite{oliver2000bayesian}. In particular, autism research could benefit from detecting asymmetry of autistic children from video (e.g., no response to name-calling \cite{liu2017response}). Another hallmark behavior of autistic children is stimming, a type of self-stimulatory behavior characterized by the repetition of physical movements such as head banging, hand flapping, and tip-toe.  

The other important hallmark for autism diagnosis is sticking to sameness \cite{gotham2013exploring} - i.e., autistic patients tend to insist on following the same routine, playing with the same toy, etc. Under the framework of ST-GCN contrastive learning, we can formulate stimming or sameness detection as a binary classification problem with different nodes on the decision tree.  Unlike the original ST-GCN design \cite{yan2018spatial}, we borrowed ideas from directed graph neural networks \cite{shi2019skeleton} and Direcformer \cite{truong2022direcformer} to replace the learnable weight matrix with a trainable difference matrix, allowing the addition of new edges. Such a warm-up strategy has shown the benefit of regularizing the model based on prior knowledge of the human body. 

\section{Experimental Results}
\label{sec:results}

\subsection{Datasets}

\textbf{UT Interaction Dataset} \cite{ryoo2009spatio}. This was one of the first datasets collected for interaction recognition. It contains six types of two-person interaction and is composed of 10 types of nonperiodic atomic-level actions. The six classes of interactions include: {\em Shake hands, point,
hug, push, kick, and punch}. The 10 types of atomic actions are {\em arm stretching, arm withdraw, leg stretching, leg lowering, 
moving forward, moving in the left and right directions}.


\noindent \textbf{NTU RGB+D.}  \cite{shahroudy2016ntu}. This was the first large-scale dataset for 3D human activity analysis. It was collected at Nanyang Technological University in 2016 from 40 distinct
subjects, containing more than 56 thousand video
samples and 4 million frames. This dataset contains 60 action classes, including daily, mutual, and health-related actions.
The authors of this dataset provided two evaluations: (A) cross-subject (CS). In this setting, clips from a selected subset of actors serve as the training set and the remaining testing set; (B) cross-view (CV). Clips from two cameras from different viewpoints are used as a training set, and the remaining clips are used as a testing set.

\noindent \textbf{NTU RGB+D 120.} \cite{liu2019ntu}. The original NTU RGB + D data set was expanded to 120 action classes, including 26 mutual actions (two-person interactions).  The video samples have been captured by three Microsoft Kinect V2 cameras simultaneously. The resolutions of the RGB videos are 1920×1080, depth maps and IR videos are all in $512\times424$, and the 3D skeletal data contains the 3D locations of 25 major body joints in each frame.
The authors also provided two evaluations similar to the original dataset: (A) cross-subject, which is the same as before; (B) cross-setup (CSet). The clips are inserted into the training and testing set according to the pre-defined setup.

\noindent \textbf{Caltech ADOS Interview.} This dataset was collected at CalTech from 2015 to 2017. It followed the ADOS-2 Module 4 protocol, consisting of 15 interview scenes. A total of 42 videos with a total duration of 3165 minutes were acquired; the average length of the ADOS interview is about 75 minutes (with a range of 50 to 150 minutes).  All videos are scored by ADOS experts based on the following five categories with 32 elements: (A) Language and Communication, (B) Reciprocal Social Interaction, (C) Imagination/Creativity, (D) Stereotyped Behaviors and Restricted Interests, and (E) Other Abnormal Behaviors. The score $0 \sim 3$ indicates the severity level of the ASD behavior targeted in that question. 0 means that the participant's response was at the level one would expect for a person without ASD, while a score of 3 would be highly indicative of ASD.

\subsection{Interaction Adjacency Matrix (IAM) and Symmetry-Related Features}
\label{Sec:5.2}

\begin{table}
    \centering
    \begin{tabular}{@{}l|cc|cc@{}}
    \toprule
    Task        & \multicolumn{2}{c|}{CS} & \multicolumn{2}{c}{CSet} \\
                & w/o IAM   & w/ IAM    & w/o IAM   & w/ IAM \\  
    \midrule 
    Asym        & 80.2\%    & \textbf{82.0}\%    & 81.3\%    & \textbf{82.7}\%    \\
    Sym         & 96.6\%    & \textbf{96.8}\%    & 96.4\%    & \textbf{96.5}\%    \\
    \hline
    Baseline    & 85.1\%    & \textbf{86.3}\%    & 86.0\%    & \textbf{87.7}\%    \\ 
    Asym + Sym  & 86.6\%    & \textbf{87.8}\%    & 87.1\%    & \textbf{88.0}\%    \\           
    \bottomrule
    \end{tabular}
    \caption{
    Comparison of the accuracy between the weighted sum of the performance of asymmetric, symmetric, and standard mutual action recognition on the mutual-actions subset of NTU RGB+D 120. ``Asym'' and ``Sym''denote the results on the subset of asymmetric and symmetric interaction, respectively.  
    } 
    \vspace{-0.3in}
    \label{tab:weighted_sum1}
\end{table}

\begin{figure}[htb]
    \centering
    \includegraphics[width=1\columnwidth]{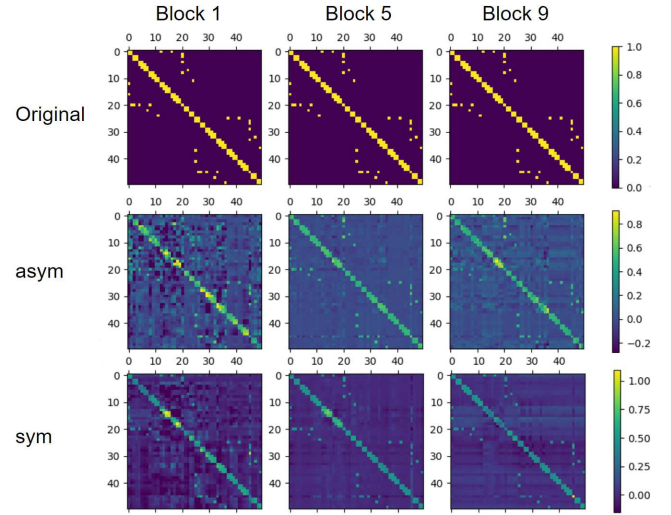}
    \caption{Visualization of the IAM learned from asym and sym mutual action recognition on NTU RGB+D 120. "Original" indicates the input IAM according to the physical connection. "asym" and "sym" denote the IAM learned from asymmetric and symmetric interaction, respectively. The columns suggest the ST-GCN blocks.}
    \vspace{-0.15in}
    \label{fig:visualize_IAM}
\end{figure}


    

\begin{figure*}
    \centering
    \begin{subfigure}[b]{0.44\textwidth}
        \includegraphics[width=\textwidth]{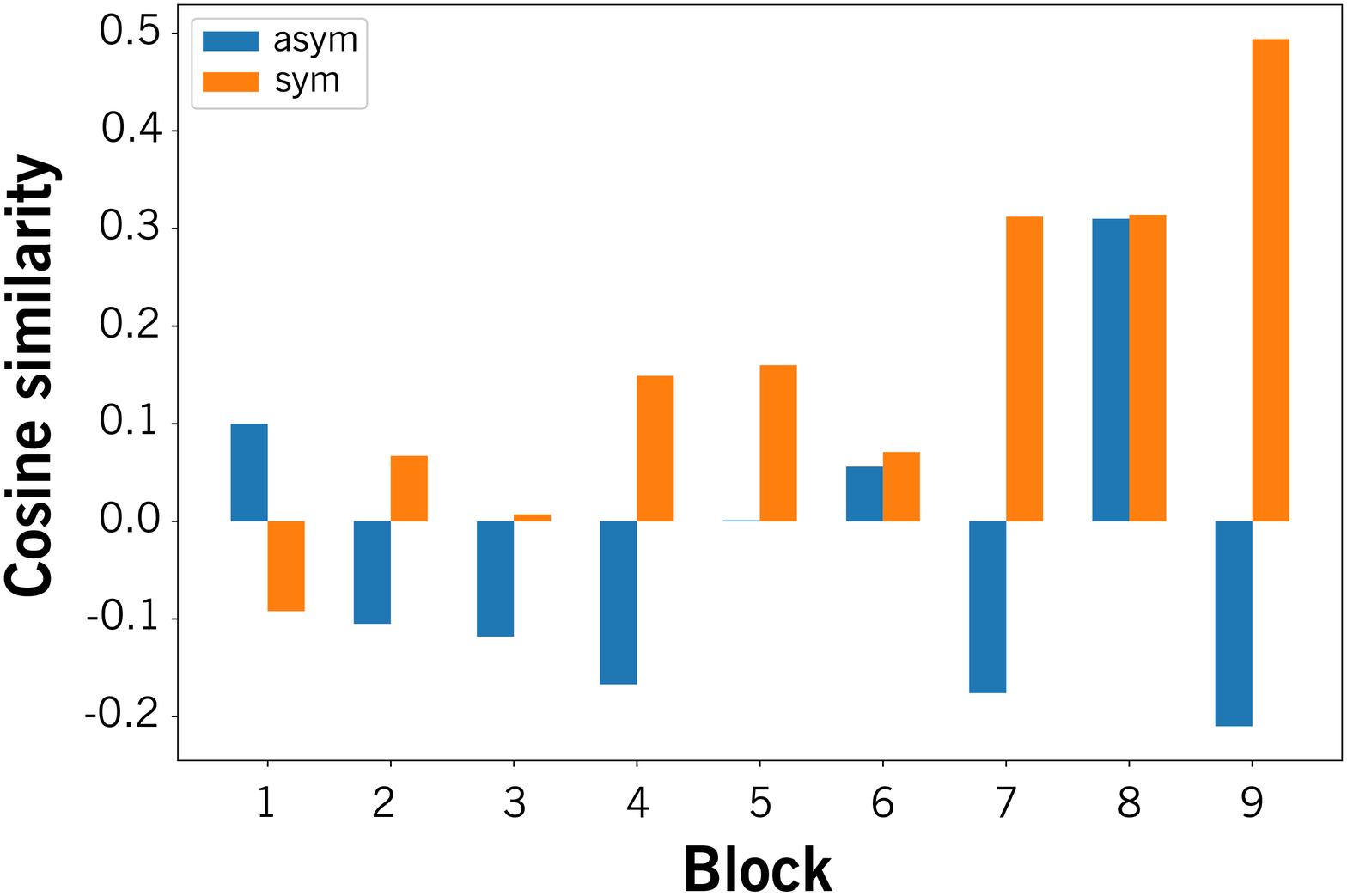}
        \caption{}
    \end{subfigure}
    \begin{subfigure}[b]{0.44\textwidth}
        \includegraphics[width=\textwidth]{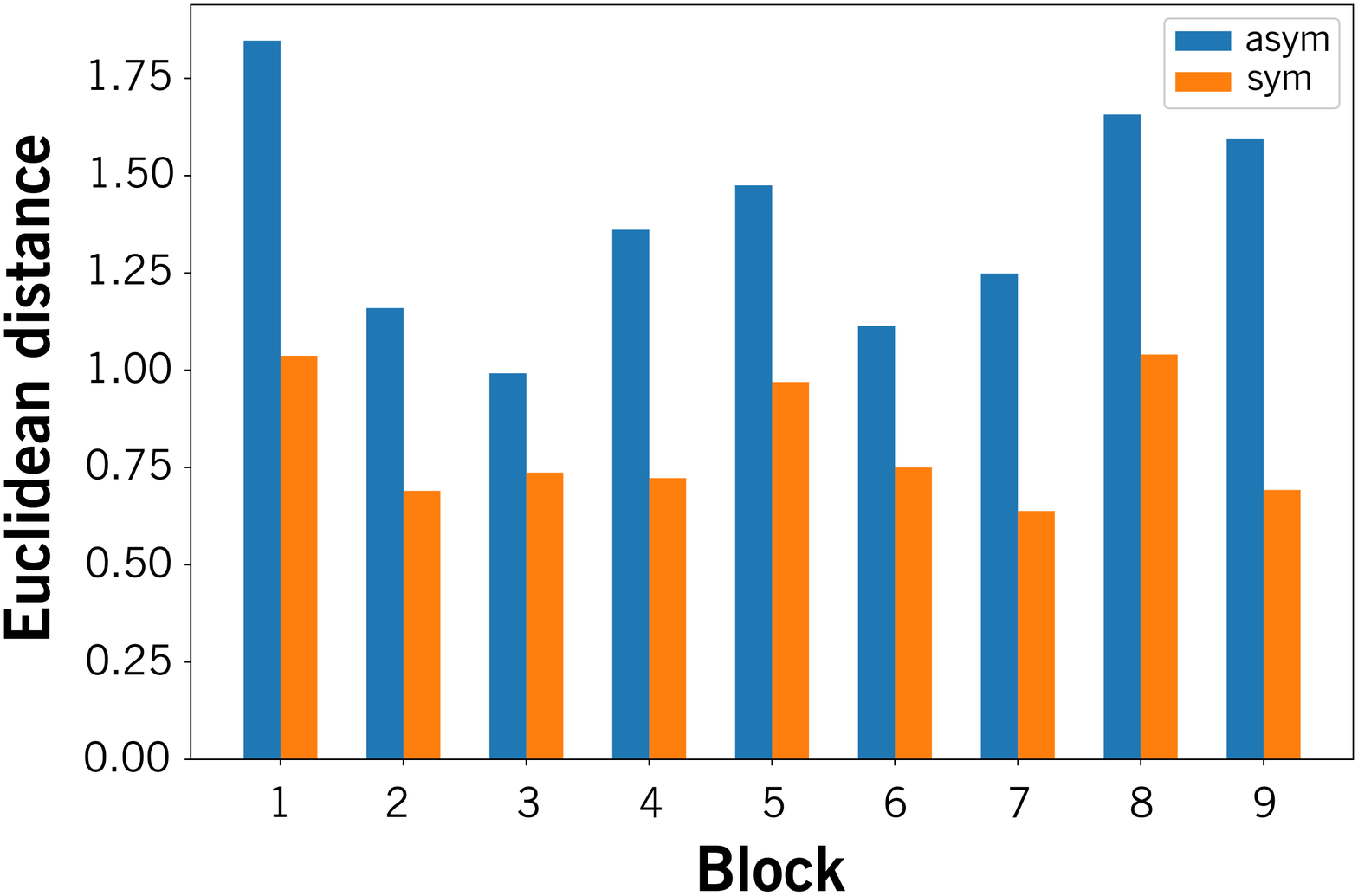}
        \caption{}
    \end{subfigure}
    \vspace{-0.1in}
    \caption{Symmetry and asymmetry comparison. (a) cosine similarity. Symmetry has a higher similarity between $\mathbf{A^{12}}$ and $\mathbf{A^{21}}$ than that of asymmetry. (b) Euclidean distance. The symmetry has a less distance between $\mathbf{A^{12}}$ and $\mathbf{A^{21}}$ than that of asymmetry.}
    \label{fig:similarity}
\end{figure*}


First, we demonstrate the benefit of learning symmetry-related features to mutual action recognition. To facilitate the study, we manually separate the mutual action subset of NTU RGB+D 120 into asym and sym classes (assuming an oracle exists) and conduct mutual action recognition experiments on them separately. ST-GCN is the backbone of this experiment, and the baseline is the standard mutual action recognition. The accuracy result is shown in Table \ref{tab:weighted_sum1}. Asym + sym indicates the weighted sum, calculated by the sum of the correct predicted sample number of asym action recognition and sym action recognition, dividing the total number of testing samples. As we can observe from Table \ref{tab:weighted_sum1}, in both CSet and CS settings, the recognition accuracy with the activated IAM module outperforms that without the activated IAM for the Asym and Sym cases, as well as the baseline and weighted sum scenarios. These results have justified the benefit of the newly designed IAM for mutual action recognition. 
Additionally, we note that the accuracy gap between Asym and Sym is as large as $15\%$, which implies that Asym is the most challenging case for mutual action recognition. Consequently, the performance gained by our IAM module can improve by 1.8\% and 1.4\% for Asym in the cross-subject and cross-setup settings, respectively. This means that our IAM module can effectively capture the asymmetric interaction between two people by off-diagonal submatrices $\mathbf{A^{12}}$ and $\mathbf{A^{21}}$.

Next, we show the feasibility of symmetry detection. As mentioned above, the symmetric interaction is supposed to have more similar characteristics between $\mathbf{A^{12}}$ and $\mathbf{A^{21}}$, which means that the edges of person 1 ($\mathbf{P_1}$) to person 2 ($\mathbf{P_2}$) are similar to the edges of $\mathbf{P_2}$ to $\mathbf{P_1}$. 
In this experiment, we initialize $\mathbf{A^{12}}$ and $\mathbf{A^{21}}$ with zero and train our model to learn IAM from the subset of asym and the subset of sym on mutual actions of NTU RGB + D 120, respectively. Visualization results of the IAM learned from asym recognition and sym recognition are shown in Figure \ref{fig:visualize_IAM}. As we can learn from the visualization, the distribution of the edge intensities in "asym" is disorganized, while those in "sym" are better aligned along the diagonal of the matrix (''asym'' appear more random than ''sym'' as shown by lighter background color and more noise-like patterns).
Furthermore, we have used cosine similarity and Euclidean distance to quantify the similarity between $\mathbf{A^{12}}$ and $\mathbf{A^{21}}$. Figure \ref{fig:similarity} shows the similarity and distance results calculated from the 9 ST-GCN blocks. It can be clearly seen from the comparison of cosine similarities that there is a striking difference between the classes of asym and sym (except for the \nth{6} and \nth{8} layers). Especially in the \nth{9} layer, the cosine similarity between asym and sym is highly distinguishable (-0.2 vs. 0.5). 
Similarly, the Euclidean distances of sym are uniformly less than those of asym, suggesting that the IAM can better preserve the similarity property of sym during training. These experimental findings have justified the feasibility of distinguishing sym from asym and have shown that IAM can separate them apart in the latent space of ST-GCN.


\begin{table}
    \centering
    \begin{tabular}{@{}lcc@{}}
    \toprule
    Method & CS & CSet \\
    \midrule
    ST-GCN\cite{yan2018spatial}     & 78.6\%            & 79.9\% \\
    AS-GCN\cite{li2019actional} & 82.9\%            & 83.7\% \\
    LSTM-IRN\cite{perez2021interaction}     & 77.7\%            & 79.6\% \\
    GeomNet\cite{GeomNet}       & \textbf{86.5}\%            & 87.6\% \\
    \hline
    Our w/o IAM                 & 85.1\%            & 86.0\% \\
    Our w/ IAM                  & 86.3\%            & \textbf{87.7}\% \\ 
    \bottomrule
    \end{tabular}
    \caption{
    Comparison of the performance on the mutual-actions subset of NTU RGB+D 120. 
    } 
    \vspace{-0.2in}
    \label{tab:sota_ntu120}
\end{table}

Finally, we show the effectiveness of IAM and symmetric-related features. Note that the objective of this work is not to advance the state-of-the-art (SOTA) in action recognition but rather to advocate the importance of understanding symmetry and dominance in mutual action recognition. 
Table \ref{tab:sota_ntu120} shows the results of our model (with and without the IAM module) and the SOTA approach on the mutual actions of NTU RGB+D 120. Our proposed IAM module can still outperform most of the competing methods. Note that our method has achieved highly competitive performance to GeomNet \cite{GeomNet} even without prior knowledge or Riemannian embedding. 

\subsection{Downstream Binary Classification Tasks}

\begin{table}
    \centering
    \begin{tabular}{@{}l | l c c@{}}
    \toprule
    Task                            & & CS & CV \\ 
    \hline 
    \multirow{ 2}{*}{UandL}       
    & upper & 83.4\% & 91.0\% \\
    & lower & 95.1\% & 97.2\% \\
    \hline 
    \multirow{ 2}{*}{Periodicity}   
    & periodic & 84.7\% & 91.8\% \\
    & aperiodic & 86.9\% & 94.3\% \\
    \midrule 
    & baseline & 85.0\% & 92.1\% \\
    Weighted sum & upper + lower & 85.6\% & 92.1\% \\
    & periodic + aperiodic & 86.1\% & 93.4\% \\
    \bottomrule
    \end{tabular}
    \caption{
    Comparison of the accuracy between the weighted sum of the performance of UandL and Periodicity binary classification tasks on the NTU RGB+D. 
    } 
    \vspace{-0.15in}
    \label{tab:weighted_sum2}
\end{table}

In this section, we first demonstrate the benefit of learning binary classification tasks, including asymmetric and symmetric (AandS), upper-body and lower-body  (UandL), and periodic and aperiodic (periodicity). And then we show the performance of our model in binary classification tasks on the NTU RGB-D dataset.
Similar to the demonstration of the benefit of learning symmetry-related features in Sec. \ref{Sec:5.2}, we manually separate the NTU RGB+D dataset into two classes (assuming that there is an oracle) depending on the binary tasks and perform action recognition experiments on them separately. ST-GCN is the backbone of this experiment, and the baseline is standard action recognition. The accuracy result is shown in Table \ref{tab:weighted_sum2}. When comparing the performance of AandS in Table \ref{tab:weighted_sum1} and the performance of UandL and Periodicity in Table \ref{tab:weighted_sum2}, we learn that Periodicity and AandS features are important for the action recognition problem, since they can improve performance by at least 1\% separately. There is less improvement from the baseline with UandL tasks since the model has a better ability to extract UandL features (see Table \ref{tab:binary}).

\begin{table}
    \centering
    \resizebox{\columnwidth}{!}{%
    \begin{tabular}{ l | c c | c c | c c | c }
        \toprule
         & \multicolumn{2}{c|}{AandS} & \multicolumn{2}{c|}{UandL} & \multicolumn{2}{c|}{Periodicity} & Dominance \\ \cline{2-8}
        Methods & \multicolumn{2}{c|}{NTU120} & \multicolumn{2}{c|}{NTU60} & \multicolumn{2}{c|}{NTU60} & UT \\ 
         & CS & CSet & CS & CV & CS & CV  \\
        \midrule
        baseline    & 92.5\% & 92.0\% & 97.3\% & 97.9\% & 90.4\% & 92.0\% & 78.8\% \\ 
        +MS2L       & 91.1\% & 91.3\% & 97.1\% & 98.1\% & 87.0\% & 91.7\% & 75.0\% \\ 
        \bottomrule
    \end{tabular}
    }
    \caption{Performance of each binary classification task. "AandS" denotes asymmetric vs. symmetric. "UandL" denotes upper vs. lower body. "periodicity" denotes periodic vs. aperiodic. "dominance" denotes dominance detection results on the UT dataset. "+MS2L" suggests a fine-tuning model (the backbone is ST-GCN) with downstream binary classification tasks.} 
    \vspace{-0.2in}
    \label{tab:binary}
\end{table}

\begin{table}
    \begin{center}
    \begin{tabular}{l c c}
    \toprule
    \multirow{2}{*}{Fold Id} & \multicolumn{2}{c}{CS}  \\
    \cline{2-3} & baseline & IAM \\
    \midrule
      1   & 100.0\% & 100.0\%  \\
      2	& 77.4\%  & 80.5\% \\
      3	& 78.3\%  & 82.6\% \\
      4	& 56.3\%  & 75.0 \% \\
      5	& 75.0\%  & 75.0\% \\
    \midrule
      Avg.	& 77.4\%   & \textbf{82.7\%}  \\
    \bottomrule
    \end{tabular}
    \end{center}
    \caption{Performance of IAM on HOI imitation scene of Caltech ADOS video dataset.}
    \vspace{-0.3in}
    \label{tab:scene8}
\end{table}

\begin{figure*}[t]
    \centering
    \includegraphics[width=2\columnwidth]{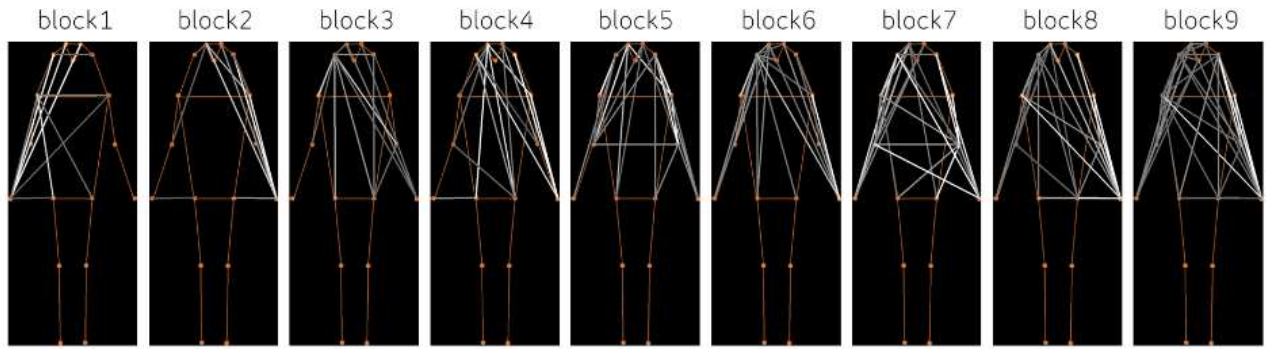}
    \caption{Visualization of significant edges in graph learned from HOI imitation on Caltech ADOS Interview dataset. The brighter the white color of the line, the larger contribution to the classification result. The orange edges indicate the initialization of the adjacency matrix according to the physical connection. The columns suggest the ST-GCN blocks.}
    \label{fig:visualization}
\end{figure*}

The results of the binary classification task are shown in Table \ref{tab:binary}, with "NTU60" and "NTU120" denoting the NTU RGB+D and NTU RGB+D 120 datasets, respectively. The evaluation strategy used is indicated by "CS" for cross-subject, "CSet" for cross-setup, and "CV" for cross-view. There are two models that we applied in this section: "baseline" (ST-GCN) and "+MS2L" (the backbone is ST-GCN). The MS2L is a self-supervised learning method. It utilizes three proxy tasks, including a motion prediction task, a jigsaw puzzle task, and a contrastive learning task, to pre-train the model and then extract the information hidden within the skeleton sequence. We fine-tuned the pre-train model with the binary classification tasks. As we can observe from Table \ref{tab:binary}, UandL achieves the best performance, as there are significant differences in spatial and temporal dimensions between upper body movements and lower body movements. Furthermore, even though MS2L is a model fine-tuned by downstream tasks, it achieved similar performance with the baseline around a 1\% accuracy gap (except for the performance on CS in the periodicity task). This suggests that the hidden information extracted by the proxy tasks of MS2L is important for binary classification.

We also tested dominance detection on the UT dataset. As shown in the tree in Figure \ref{fig:decision}, the detection of dominance is the follow-up task for asym. We have applied ST-GCN and our contrastive learning model for this task. Due to the relatively small size of the UT-interaction dataset, we combine UT-1 and UT-2 for dominance detection. There are four types of asym (kicking, pointing, punching, and pushing) in the UT-interaction dataset. Table \ref{tab:binary} shows the performance of our model for dominance detection.

\begin{figure*}[t]
    \centering
    \includegraphics[width=2\columnwidth]{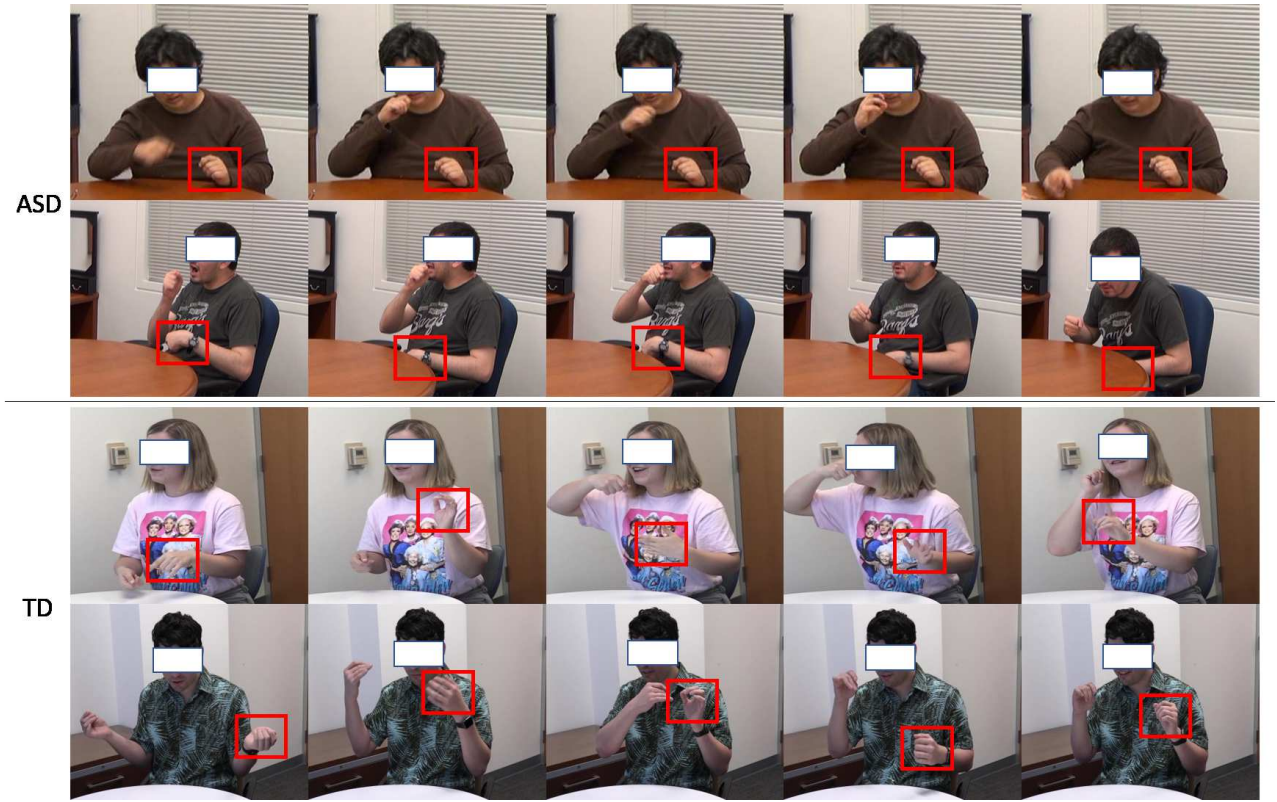}
    \caption{Example of gesture differences between ASD and TD of HOI imitation on Caltech ADOS Interview dataset.}
    \label{fig:ADOS_examples}
\end{figure*}

\subsection{ASD Diagnosis on Caltech ADOS Interview}
\label{sec:ADOS}
To verify the effectiveness of our model, we applied it to HOI imitation task, which is one of the 15 interview scenes in the Caltech ADOS Interview dataset \cite{zhang2022discriminative}. Specifically, the participants are asked to demonstrate and describe how they brush their teeth. We pre-train the model using the NTU RGB+D 120 dataset and fine-tune it with the HOI imitation task data. Five-fold cross-validation is applied to estimate the performance of our model on the ASD diagnosis (ASD vs. typical development binary classification) based on atypical motor behavior, shown in Table \ref{tab:scene8}. 

To explain the performance of the action recognition model in ASD diagnosis, we visualize nine adaptive graphs (each from an ST-GCN block) learned from the HOI imitation task in Figure \ref{fig:visualization}. The graphs show which edges are significant and contribute to the classification. The brighter the line, the larger contribution to the classification. A joint with a higher degree of bright edges is more significant for the classification. In the HOI imitation task, right-handed participants use their right hand to pretend they are holding a toothbrush. These low-level features of the right-hand movement are extracted by the shallowest ST-GCN block (block1). As the block goes deeper and the receptive field becomes wider, the joint of the left hand becomes more significant (block2 to block6). Finally, both left and right hands are taken into account, and the left hand is the most significant feature in the deepest layer (block 9). These visualization results are strong evidence showing that the left-hand feature plays a significant role in the ASD diagnosis of a right-handed person. 
As shown in Figure \ref{fig:ADOS_examples}, due to the atypical motor behavior, the participants with ASD usually avoid expressive gestures on their left hand, like gripping their fists. In sharp contrast, participants with typical development usually use express gestures on their left hand to describe how they brush their teeth. We have verified that this is indeed the most salient feature used by clinicians while inspecting the task of HOI imitation. By extracting symmetric features, the introduction of IAM can avoid the bias between left-handed and right-handed individuals and outperform the baseline.
\section{Conclusion}
\label{sec:conclude}
In this paper, we propose a decision tree-based contrastive learning framework for behavioral imaging and demonstrate its applications in action recognition and diagnosis of ASD. Using a skeleton graph as the data representation, we have constructed an interaction adjacent matrix on the graph convolutional network as the backbone for modeling action-related attributes.  Through the construction of various
pretext tasks and loss functions, we have designed a series of binary classification nodes on the decision tree, which can be combined to support higher-level vision tasks such as action recognition and ASD diagnosis. For action recognition, we have focused on the experiments with periodicity and dominance attributes; for ASD diagnosis, we have studied symmetry and sameness attributes. 
Our preliminary experimental results have shown the promising performance of the decision tree-based contrastive learning approach on four binary attribute classifications (AandS, UandL, Periodicity, and Dominance) on UT Interaction, NTU-60, and NTU-120 datasets. For ASD diagnosis, our approach has achieved a promising $>80\%$ accuracy on the challenging CalTech ADOS Interview database. In addition to good accuracy, transparency of our ST-GCN based approach is another desirable property because interpretability matters in ASD-related clinical practice.
However, the effectiveness of our decision tree-based contrastive learning method is contingent upon the quality and quantity of binary attributes. As we continue our research, future work includes the expansion of constructed decision trees (i.e., extraction of more action-related and ASD-related binary attributes), optimization of contrastive learning loss functions and hyper-parameters, and attention-based fusion of multiple nodes to support multi-label action recognition and fine-granularity diagnosis of ASD.

\vspace{-0.1in}
\begin{center}
    \textbf{Acknowledgement}
\end{center}
The authors thank anonymous reviewers for constructive comments that help improve the presentation of this paper. This work is partially supported by NSF awards BCS-1945230 and HCC-2114644.

\bibliographystyle{ACM-Reference-Format}
\bibliography{6_ref}

\end{document}